\documentclass[letterpaper]{article} 
\usepackage{aaai2026}  
\usepackage{times}  
\usepackage{helvet}  
\usepackage{courier}  
\usepackage[hyphens]{url}  
\usepackage{graphicx} 
\usepackage{bm}
\usepackage{threeparttable}
\usepackage{longtable}
\usepackage{booktabs}
\usepackage{multirow}
\usepackage{amssymb} 
\usepackage{mathtools}
\usepackage{natbib}  
\usepackage{caption} 
\usepackage{algorithm}
\usepackage{algorithmic}

\usepackage{algorithm}
\usepackage{algorithmic}
\usepackage[table]{xcolor}

\usepackage{newfloat}
\usepackage{listings}
\DeclareCaptionStyle{ruled}{labelfont=normalfont,labelsep=colon,strut=off} 
\floatstyle{ruled}
\newfloat{listing}{tb}{lst}{}
\floatname{listing}{Listing}
\title{What-Meets-Where: Unified Learning of Action and Contact Localization in Images}
\author{
    Yuxiao Wang\textsuperscript{\rm 1}, Yu Lei\textsuperscript{\rm 2}, Wolin Liang\textsuperscript{\rm 1}, Weiying Xue\textsuperscript{\rm 1}, Zhenao Wei\textsuperscript{\rm 3}, Nan Zhuang\textsuperscript{\rm 4}, Qi Liu\textsuperscript{\rm 1}\thanks{Corresponding author}
}
\affiliations{
    \textsuperscript{\rm 1}School of Future Technology, South China University of Technology\\
    \textsuperscript{\rm 2}School of Information Science \& Technology, Southwest Jiaotong University\\
    \textsuperscript{\rm 3}Computing Science and Artificial Intelligence College, Suzhou City University \\
    \textsuperscript{\rm 4}School of Software Technology, Zhejiang University

    ftwangyuxiao@mail.scut.edu.cn, drliuqi@scut.edu.cn
}

\usepackage{bibentry}

\begin{document}

\maketitle

\begin{figure*}[!h]
\centering 
\includegraphics[width=\linewidth]{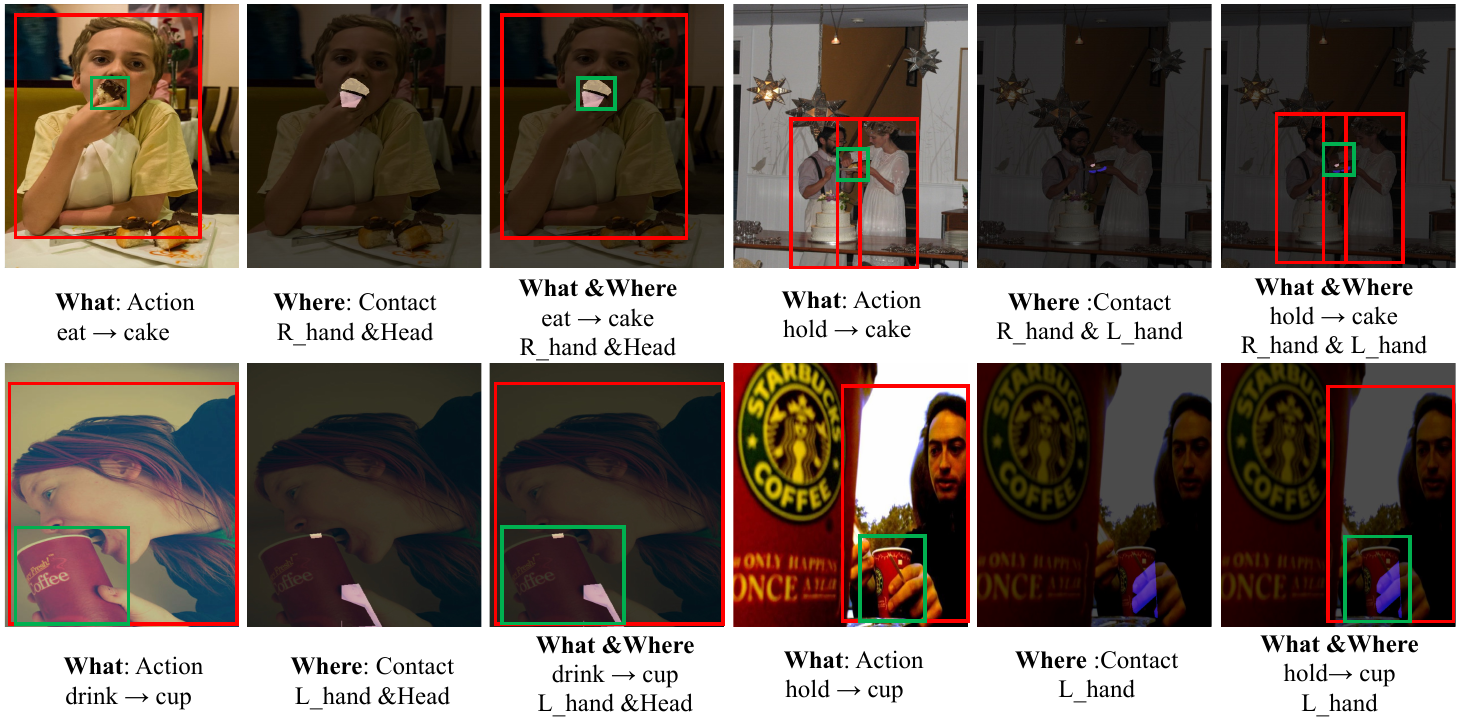}
\caption{The same object (e.g., cake or cup) can imply different actions depending on contact regions. For example, ``eating'' involves both hand and head, while ``holding'' involves only the hand. To bridge this gap, we propose joint modeling of \textbf{What} (action \& object) and \textbf{Where} (contact body part).}
\label{fig:inn}
\end{figure*}
\begin{abstract}
People control their bodies to establish contact with the environment. To comprehensively understand actions across diverse visual contexts, it is essential to simultaneously consider \textbf{what} action is occurring and \textbf{where} it is happening. Current methodologies, however, often inadequately capture this duality, typically failing to jointly model both action semantics and their spatial contextualization within scenes. To bridge this gap, we introduce a novel vision task that simultaneously predicts high-level action semantics and fine-grained body-part contact regions. Our proposed framework, PaIR-Net, comprises three key components: the Contact Prior Aware Module (CPAM) for identifying contact-relevant body parts, the Prior-Guided Concat Segmenter (PGCS) for pixel-wise contact segmentation, and the Interaction Inference Module (IIM) responsible for integrating global interaction relationships. To facilitate this task, we present PaIR (Part-aware Interaction Representation), a comprehensive dataset containing 13,979 images that encompass 654 actions, 80 object categories, and 17 body parts. Experimental evaluation demonstrates that PaIR-Net significantly outperforms baseline approaches, while ablation studies confirm the efficacy of each architectural component.

\end{abstract}

\begin{links}
    \link{Code}{https://drliuqi.github.io/}
    \link{Extended version}{https://arxiv.org/abs/2508.09428}
\end{links}

\section{Introduction}
Individuals interact with the environment through both behavioral intent and physical contact, whether sitting on a chair or lifting a cup. This dual understanding, combining semantic goals with physical contact mechanisms, is fundamental to human actions. Neuroscience supports this view, showing that actions like ``pushing a door'' involve not only action recognition but also verification of contact between the hand and the door handle~\cite{bicchi2000robotic}.

However, existing benchmarks fail to capture both aspects simultaneously. While prior works focus on specific body-part contact (e.g., hand-object~\cite{shan2020understanding,cui2023hand}, foot-ground~\cite{tripathi20233d}, or human-object contact~\cite{hot}), they are often limited to localized contexts and lack full-body interaction modeling. In contrast, action recognition methods typically emphasize ``\textbf{what}'' action occurs, neglecting the ``\textbf{where}'' of physical contact, thus hindering their applicability in complex, real-world scenarios~\cite{wang2025precision, wang2024freea, wang2024dehot}.

As shown in Figure~\ref{fig:inn}, a single object (e.g., a cake or cup) can correspond to different actions based on contact regions. For instance, ``eating'' involves the hand and head, while ``holding'' requires only the hand. This underscores the need to jointly model \textbf{what} the action is and \textbf{where} contact occurs. Beyond action classification~\cite{wang2022general}, models should also localize body parts involved (e.g., ``buttocks contacting chair'' for "sitting"), essential for fine-grained understanding~\cite{wang2025precision} and applications like robotic planning, AR/VR assessment and imitation learning~\cite{westermeier2024assessing, zare2024survey}.

To address this, we introduce a new task that jointly infers action semantics and physical contact segmentation. The task poses two interconnected questions: \textbf{What} is the person doing, and \textbf{where} is the body in contact with the object to perform the action? This requires a dataset that captures both high-level intent and low-level grounding. \textbf{We propose \textbf{PaIR} (Part-aware Interaction Representation), a dataset of 13,979 real-world images with 654 action types, 80 object categories, and 17 contactable body parts.} Each sample includes annotations of $\langle$\textit{person}, \textit{verb}, \textit{object}, \textit{contact part}$\rangle$, 2D contact masks, and part labels, enabling joint learning of interaction semantics and contact.

We propose PaIR-Net, a unified framework that jointly models action recognition and contact segmentation. PaIR-Net consists of three key components: the Contact Prior Aware Module (CPAM) predicts interacting body parts; the Prior-Guided Concat Segmenter (PGCS) segments contact regions; and the Interaction Inference Module (IIM) integrates spatial and semantic cues to infer interaction types. Additionally, two mechanisms enhance performance: the H-O RoI Enhancer for guiding PGCS with bounding boxes and the Mask-Guided RoI Feature module leverages contact masks for better recognition. 

\section{Related Works}
\textbf{Action Recognition}.
Standard object detectors~\cite{dai2016r,ren2015faster} localize instances but fail to capture inter-instance interactions. Early works address this using pose cues~\cite{desai2012detecting,yao2010modeling}, while visual phrase models~\cite{sadeghi2011recognition} detect interacting pairs. Recent human-object interaction methods~\cite{liao2022gen,yang2024open} improve fine-grained recognition by identifying specific human-object pairs, and others~\cite{wang2024ted,wang2024cyclehoi} incorporate context and language priors.

\textbf{Contact Perception}.
Recent studies explore contact modeling at specific body parts, such as hands~\cite{shan2020understanding, darkhalil2022epic} and feet~\cite{tripathi20233d}, revealing the value of contact-aware understanding. However, these methods often focus on isolated parts or rely on specific modalities like depth or video. Although human-object contact~\cite{hot} introduces a full-body contact dataset to detect multiple contact points, it lacks integration with semantic action reasoning.
In contrast, our approach unifies semantic action recognition and dense contact localization from a single 2D image.

\textbf{Datasets}.
Several recent datasets focus on physical contact between the human body and the environment. VISOR~\cite{darkhalil2022epic} serves as a dataset for pixel-level annotations and a benchmark for segmenting hands~\cite{ma2022hand} and active objects in egocentric video~\cite{girdhar2019video}. PressurePose~\cite{clever2020bodies} provides 206K synthetic pressure images with paired 3D pose and shape. RICH~\cite{huang2022capturing} captures real-world multi-view interactions in 3D scenes, while ContactPose~\cite{brahmbhatt2020contactpose} and ARCTIC~\cite{fan2023arctic} focus on hand or whole-body contact. HOT~\cite{chen2023detecting} enables contact detection from images, yet these datasets primarily emphasize contact perception and lack integration with action semantics~\cite{wang2025prompt}. Conversely, action recognition datasets such as imSitu~\cite{yatskar2016situation}, HICO~\cite{chao2015hico}, HICO-Det~\cite{chao2018learning}, and V-COCO~\cite{gupta2015visual} provide $\langle$human, verb, object$\rangle$ triplets but omit contact-specific annotations. To bridge this gap, we introduce PaIR, the first dataset to provide $\langle$\textit{human}, \textit{verb}, \textit{object}, \textit{contact part}$\rangle$ annotations at the image level. 

\begin{figure*}[t]
\centering 
\includegraphics[width=\linewidth]{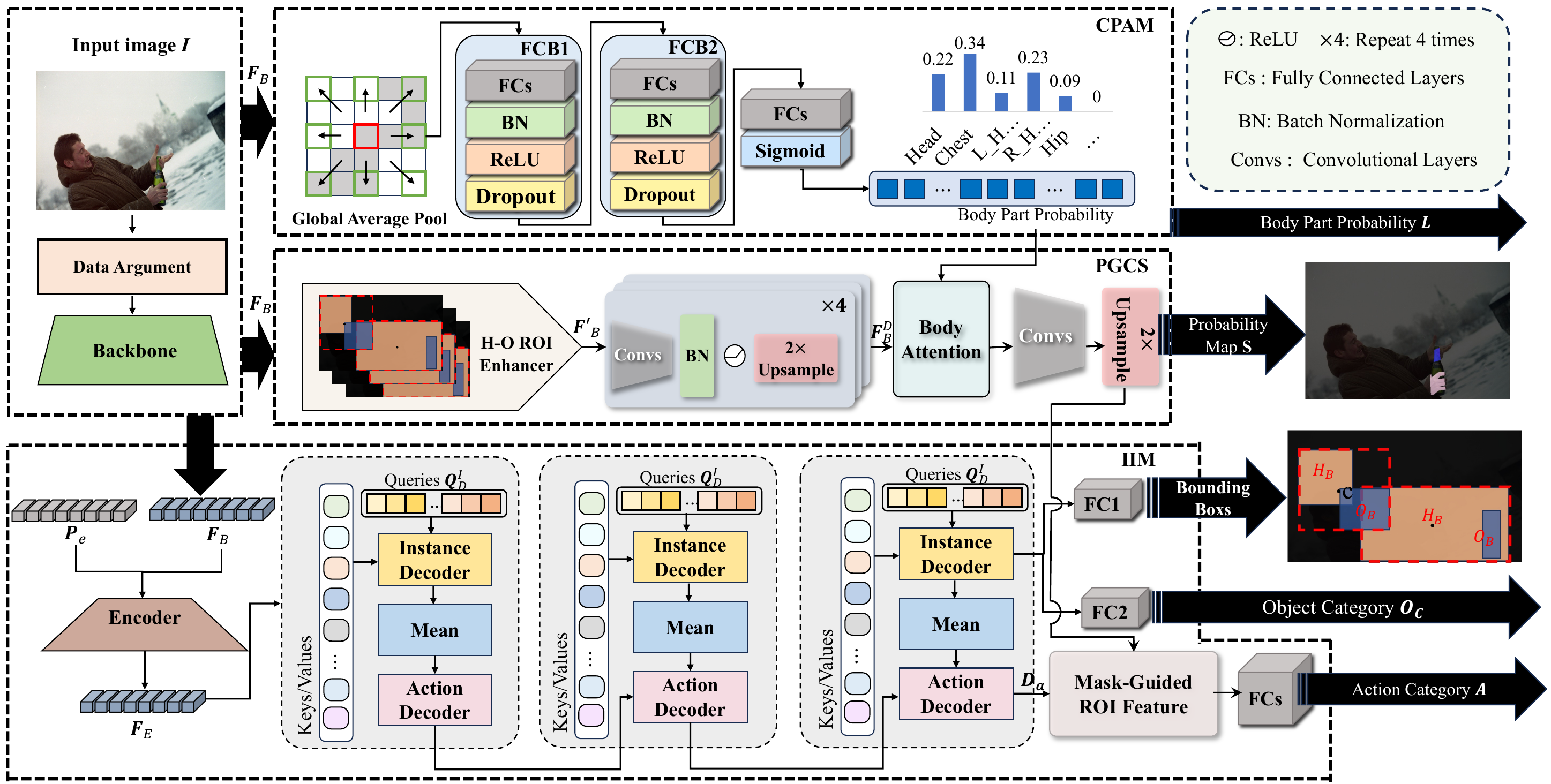}
\caption{The overall workflow of PaIR-Net. It comprises three branches: CPAM for multi-label body part contact prediction (upper part of Figure), PGCS for outputting contact region segmentation (middle part of Figure), and IIM for detecting human-object pairs and identifying interaction categories (lower part of Figure). To facilitate effective collaboration between contact understanding and action recognition, we design two key modules: the H-O RoI Enhancer and the Mask-Guided RoI Feature.}
\label{fig:network}
\end{figure*}

\section{Method}
In this paper, we propose a novel framework, PaIR-Net, to jointly model action semantics and contact regions. 
Specifically, PaIR-Net can identify which human-object pairs are interacting in an image, classify the type of interaction, and segment the specific body parts involved in the contact.
\subsection{Overall Architecture}
The architecture of PaIR-Net is shown in Figure \ref{fig:network}. For input image $\bm{I} \in \mathbb{R}^{H\times W\times 3}$, we apply data augmentation techniques \cite{wang2024ted} before using ResNet \cite{he2016deep}, or Swin Transformer \cite{liu2021swin} as backbone to extract feature $\bm{F}_B \in \mathbb{R}^{\frac{H}{s}\times \frac{W}{s} \times C}$, where $s=32$ and $C$ is the channel dimension. Then a CPAM network is employed to focus on 17 body parts that contact objects. The $\bm{F}_B$ also feeds into both the PGCS and the IIM to generate contact region segmentation and interaction actions, respectively. For effective collaboration between these modules, we introduce the H-O RoI Enhancer, which guides PGCS to focus on potential interaction areas, and the Mask-Guided RoI Feature module, which leverages the segmentation results of PGCS to enhance interaction recognition.

\subsection{Contact Prior Aware Module (CPAM)}

To guide PGCS toward interaction-relevant regions, CPAM is designed. The module first applies global average pooling (GAP) to $\bm{F}_B$, ensuring consistent dimensions for subsequent processing. The pooled features pass through fully connected blocks (FCB)—each containing a Fully Connected (FC) layer, Batch Normalization (BN), ReLU, and Dropout—to enhance semantic representation. Then, a sigmoid activation is used to output the contact probability for each body part, helping PGCS focus on body parts likely involved in interactions. Specifically, given features $\bm{F}_B \in \mathbb{R}^{\frac{H}{s}\times \frac{W}{s} \times C}$, we apply GAP to obtain $\bm{F}_G \in \mathbb{R}^{1\times 1 \times C}$, which passes through two FCB to produce the final contact prediction $\bm{L} \in \mathbb{R}^{N_c}$, where $N_c$ represents the number of contact categories. The process can be formulated as:
\begin{equation}
\bm{L} = \text{Sigmoid}(\text{FC}\left(\text{FCB}_2\left(\text{FCB}_1\left(\text{GAP}(\bm{F}_B)\right)\right)\right)),
\end{equation}
where $\text{FCB}_i(\cdot)$ represents the $i$-th fully connected block. Since the CPAM performs a multi-label classification task, $\bm{L}$ is compared with the ground-truth contact labels using the Binary Cross-Entropy (BCE) loss.

\begin{figure*}[h]
\centering 
\includegraphics[width=\linewidth]{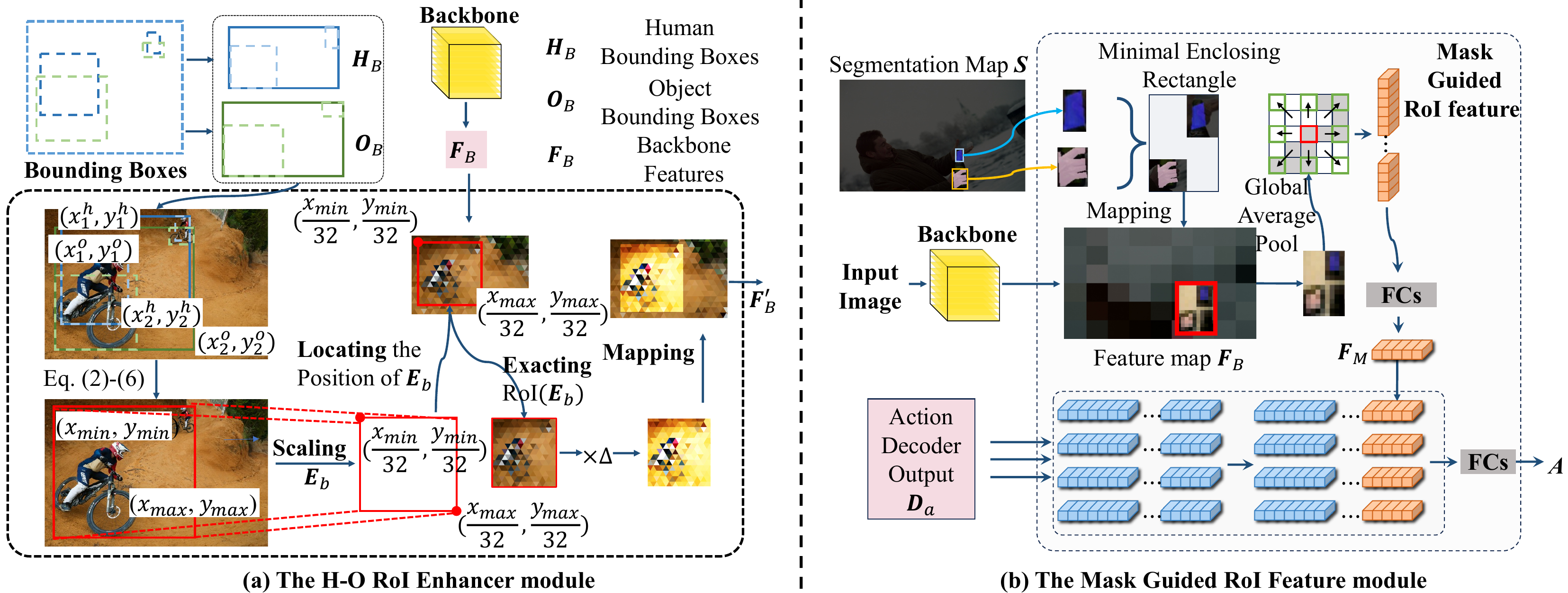}
\caption{(a) The H-O RoI Enhancer module. It computes the minimum enclosing rectangle based on the human and object bounding boxes, and enhances the feature $\bm{F}_B$ responses within this region. (b) The structure of the Mask-Guided RoI Feature module. It utilizes $\bm{S}$ to extract the minimum enclosing contact region,  crops the corresponding region from $ \bm{F}_B $, and generates the contact feature encoding $ \bm{F}_M $ through GAP and FC layers. Finally, $ \bm{F}_M $ is fused with $ \bm{D}_a $ to assist action classification.
}
\label{fig:two_module}
\end{figure*}

\subsection{Prior-Guided Concat Segmenter (PGCS)}

The PGCS produces pixel-level segmentation maps of contact regions through two key submodules: the H-O ROI Enhancer Module that refines region-level features, and the Body Attention mechanism that provides body part relevance priors for improved contact localization.

\textbf{H-O RoI Enhancer Module}. Since physical contact is inseparable from interaction areas, inspired by this, we propose the H-O RoI Enhancer module to enhance features of potential contact regions, guiding the PGCS network to focus on interaction-related areas. As shown in Figure \ref{fig:two_module}(a), the module takes feature map $\bm{F}_B$, human bounding boxes $\bm{H}_B = [\bm{h}_{b1}, \bm{h}_{b2}, \ldots, \bm{h}_{bn}]$ and object bounding boxes $\bm{O}_B = [\bm{o}_{b1}, \bm{o}_{b2}, \ldots, \bm{o}_{bn}]$ predicted by the IIM module. Each bounding box $\bm{h}_{b_i}$ or $\bm{o}_{b_i}$ is defined by $(x_1, y_1, x_2, y_2)$, representing the top-left and bottom-right coordinates. For each pair of the sets of $\bm{H}_B$ and $ \bm{O}_B$, we compute the minimum enclosing rectangle $\bm{E}_b=(x_{\text{min}}, y_{\text{min}}, x_{\text{max}}, y_{\text{max}})$ to determine the feature region for enhancement, via:
\begin{equation}
    x_{\min}
    = \min\!\left\{
        \min_{1 \le i \le n} x_1(\bm{h}_{b_i}),
        \;
        \min_{1 \le j \le m} x_1(\bm{o}_{b_j})
    \right\},
\end{equation}
\begin{equation}
    y_{\min}
    = \min\!\left\{
        \min_{1 \le i \le n} y_1(\bm{h}_{b_i}),
        \;
        \min_{1 \le j \le m} y_1(\bm{o}_{b_j})
    \right\},
\end{equation}
\begin{equation}
    x_{\max}
    = \max\!\left\{
        \max_{1 \le i \le n} x_2(\bm{h}_{b_i}),
        \;
        \max_{1 \le j \le m} x_2(\bm{o}_{b_j})
    \right\},
\end{equation}
\begin{equation}
    y_{\max}
    = \max\!\left\{
        \max_{1 \le i \le n} y_2(\bm{h}_{b_i}),
        \;
        \max_{1 \le j \le m} y_2(\bm{o}_{b_j})
    \right\},
\end{equation}
\begin{equation}
    \bm{E}_b = (x_{\text{min}}, y_{\text{min}}, x_{\text{max}}, y_{\text{max}}),
\end{equation}
where $n$ and $m$ represent the number of human and object bounding boxes, respectively. Functions $x_1(\cdot)$ and $y_1(\cdot)$ extract the x and y coordinates of the top-left corner, while $x_2(\cdot)$ and $y_2(\cdot)$ extract the bottom-right corner coordinates. 
The resulting $\bm{E}_b$ is the minimal enclosure rectangle that completely covers all the human and object bounding boxes. To ensure spatial correspondence between RoI regions $\bm{E}_b$ and backbone features $\bm{F}_B$, $\mathbf{E}_b$ is downscaled by a factor of 32.
Subsequently, we amplify the feature values within this region by a factor of $\Delta$, thereby enhancing the PGCS's attention to the instance interaction areas, via:
\begin{equation}
    \bm{F}_B'(i, j) = 
\begin{cases}
\Delta \times \bm{F}_B(i, j), & \text{if } (i, j) \in \text{RoI}(\bm{E}_B) \\
\bm{F}_B(i, j), & \text{otherwise}
\end{cases},
\end{equation}
where $\text{RoI}(\bm{E}_B)$ denotes the set of all pixel coordinates inside the scaled bounding box $ \bm{E}_B $. $\Delta$ is a learnable parameter, which is initialized to 1.0.

Subsequently, the $\bm{F}_B'$ are fed into a multi-stage decoder consisting of convolutional layers, BN, ReLU activation functions, and $2\times$ upsampling layers. As a result, the decoded features $\bm{F}_B^D \in \mathbb{R}^{\frac{H}{2} \times \frac{W}{2} \times 64}$ are obtained.

\textbf{Body Attention}. Given the $\bm{F}_B^D$ and the $\bm{L} \in \mathbb{R}^{N_c}$ (predicted by the CPAM module), the body attention mechanism enhances the features through the following steps:
\begin{equation}
\bm{L}' = \text{Sigmoid}(\text{FC}(\text{ReLU}(\text{FC}(\bm{L})))),
\end{equation}
\begin{equation}
\bm{F}_B^{D'} = \bm{F}_B^D \odot \bm{L}'.
\end{equation}

The two $\text{FC}$ layers expand the feature representation to obtain $\bm{L}' \in \mathbb{R}^{1\times1\times64}$. $\odot$ denotes element-wise multiplication along the channel dimension. $\bm{F}_B^{D'}$ will pass through a convolutional layer followed by a 2× upsampling layer, producing the segmentation probability map $\bm{S} \in \mathbb{R}^{H\times W\times (N_c+1)}$ for contact regions, where the additional dimension represents the background.

\begin{table*}[ht]
\centering
\setlength{\tabcolsep}{1.0mm}{
\begin{tabular}{llcl|ccccc}
\specialrule{1.5pt}{0pt}{0pt} 
\multirow{2}{*}{Methods}                                                                                                       & \multirow{2}{*}{Params} & \multirow{2}{*}{Times (ms)} & \multirow{2}{*}{Backbone} & \multicolumn{1}{c}{\multirow{2}{*}{mAP$\uparrow$}} & \multirow{2}{*}{SC-Acc.$\uparrow$} & \multirow{2}{*}{C-Acc.$\uparrow$} & \multirow{2}{*}{mIoU$\uparrow$} & \multirow{2}{*}{wIoU$\uparrow$} \\
                                                                                                         &       &               &                           & \multicolumn{1}{c}{}                     &                          &                         &                       &                       \\ \hline
Trans~\shortcite{tamura2021qpic}+UNet~\shortcite{ronneberger2015u}         & 74.0M  & 88.7          & ResNet-50                 & 13.63                                     & 14.80                    & 26.66                   & 9.33                  & 15.10                 \\
RCLTrans~\shortcite{kim2023relational}+UNet~\shortcite{ronneberger2015u}          & 107.7M   & 123.8  & ResNet-50                 & 24.80                                    & 17.59                    & 33.43                   & 11.63                 & 15.92                 \\
STTrans~\shortcite{zhang2022exploring}+UNet~\shortcite{ronneberger2015u} & 87.2M & 114.9 & ResNet-50                 & 28.48                                    & 15.04                    & 27.16                   & 9.37                  & 13.04                 \\

DisTrans~\shortcite{wang2024ted}+UNet~\shortcite{ronneberger2015u}          & 79.8M        & 102.3                                            & ResNet-50                 & 31.23                                    & 16.67                    & 30.58                   & 11.39                 & 15.58                 \\
DisTrans~\shortcite{wang2024ted}+LinkNet~\shortcite{chaurasia2017linknet}   & 120.4M    & 83.9                                                & ResNet-50                 & 31.17                                    & 20.34                    & 35.79                   & 11.78                 & 15.51                 \\
DisTrans~\shortcite{wang2024ted}+MANet~\shortcite{fan2020ma}   & 236.6M     & 89.5                                                            & ResNet-50                 & 31.01                                    & 22.91                    & 44.32                   & 12.72                 & 17.36                 \\
DisTrans~\shortcite{wang2024ted}+HOT~\shortcite{hot}       & 97.5M   & 230.4                                                                 & ResNet-50                 & 31.19                                    &   23.24 & 38.72 & 13.30 & 18.00                      \\
DisTrans~\shortcite{wang2024ted}+PIHOT~\shortcite{wang2025precision}  & 158.6M  & 257.5                                                                      & ResNet-50                 & 31.45                                    & 26.35 & 45.16 & 14.53 & 18.54                       \\
\rowcolor[gray]{0.9}Ours                            & \textbf{55.6M}   & \textbf{75.2}                                                                                           & ResNet-50                 & 35.09                                    & 29.93                    & 50.87                   & 17.79                 & 22.23                 \\
\rowcolor[gray]{0.9}Ours                    & 74.6M     & 91.4                                                                                                  & ResNet-101                & 36.24                                    & 31.01        & 52.35               & 18.89                      & 22.59                                           \\ 
\rowcolor[gray]{0.9}Ours                   & 73.6M   & 74.3                                                                                                     & Swin-S                &36.85                                    & 33.47         &53.83               &19.33                    & 23.04                                          \\ 
\rowcolor[gray]{0.9}Ours               & 224.2M    & 119.6                                                                                                        & Swin-L                & \textbf{38.07}                                    & \textbf{37.14}         & \textbf{56.98}               & \textbf{20.95}                      & \textbf{24.12}                                           \\ 
\specialrule{1.5pt}{0pt}{0pt} 
\end{tabular}}

\caption{Performance comparisons on PaIR-1 dataset. }
\label{tab:pair-1}
\end{table*}

\subsection{Interaction Inference Module (IIM)}
\label{sec_ar}

To comprehensively infer interactions, we design an IIM based on an encoder-decoder architecture. In the encoder, parameters are initialized using DETR~\cite{carion2020end,liao2022gen}. Specifically, $\bm{F}_B$ is first embedded with positional encoding $\bm{P}_e$, then processed by the encoder to produce encoded features $\bm{F}_E$:
\begin{equation}
\bm{F}_E = \text{Encoder}(\bm{F}_B + \bm{P}_e).
\end{equation}

Subsequently, we initialize two query matrixes for humans and objects: $\bm{Q}_h\in\mathbb{R}^{N_q\times C_Q}$ and $\bm{Q}_o\in\mathbb{R}^{N_q\times C_Q}$, where $C_Q$ is the feature dimension and $N_q$ is the maximum number of instances. These are concatenated into a single query set $\bm{Q}_D^I \in \mathbb{R}^{2N_q \times C_Q}$, which will input to the instance decoder, with $\bm{F}_E$ serving as both key $\bm{K}_D^I$ and value $\bm{V}_D^I$. The instance decoder outputs instance features $\bm{D}_i$, which are split into human features $\bm{D}_h$ and object features $\bm{D}_o$:
\begin{equation}
\bm{D}_i = \text{InstanceDecoder}(\bm{Q}_D^I, \bm{K}_D^I, \bm{V}_D^I),
\end{equation}
\begin{equation}
    \bm{D}_h, \bm{D}_o = \text{Split}(\bm{D}_i),
\end{equation}
where $\bm{D}_h, \bm{D}_o \in \mathbb{R}^{N_q \times C_q}$, and $\text{Split}(\cdot)$ divides the input matrix into two parts along the row dimension. To understand interaction between each human-object pair, we add$\bm{D}_h$ and $\bm{D}_o$ and average them to obtain the query vector $\bm{Q}_D^A$. This query feeds into the Action Decoder, with $\bm{F}_E$ serving as both key ($\bm{K}_D^A$) and value ($\bm{V}_D^A$). The action decoder then outputs action features $\bm{D}_a$.
\begin{equation}
\bm{D}_a = \text{ActionDecoder}(\frac{\bm{D}_h + \bm{D}_o}{2}, \bm{K}_D^A, \bm{V}_D^A).
\end{equation}

After three consecutive decoder stages, we obtain new $\bm{D}_h$, $\bm{D}_o$, and $\bm{D}_a$. $\bm{D}_h$ and $\bm{D}_o$ pass through an FC1 network to predict human bounding boxes $\bm{H}_B \in \mathbb{R}^{N_q \times 4}$ and object bounding boxes $\bm{O}_B \in \mathbb{R}^{N_q \times 4}$, respectively. Additionally, $\bm{D}_o$ feeds into an FC2 network to predict object category $\bm{O}_C \in \mathbb{R}^{N_q}$. $\bm{D}_a$ are input to our mask-guided RoI feature module, which integrates the segmentation map $\bm{S}$ from PGCS to predict action categories $\bm{A}\in \mathbb{R}^{N_q}$.

\textbf{Mask-guided RoI Feature Module}. Contact regions between humans and objects provide key action recognition cues—sitting involves buttocks-chair contact, while holding requires hand-bottle contact. Therefore, the segmentation map $\bm{S}$ assists action classification. As shown in Figure \ref{fig:two_module}(b), we extract all non-background contact regions from $\bm{S}$ and compute their bounding boxes. The minimal enclosing rectangle is determined, and the corresponding sub-region is cropped from $\bm{F}_B$ to obtain focused interaction features. We then apply GAP followed by a fully-connected layer to generate a compact feature $\bm{F}_M \in \mathbb{R}^{10}$, which serves as the encoded representation of contact regions.
\begin{equation}
    \bm{F}_M = \text{FC}(\text{GAP}(\text{Crop}(\bm{F}_B, S))).
\end{equation}

$\bm{F}_M$ and $\bm{D}_a$ are concatenated and passed through FC layers to output the final action category predictions $\bm{A}\in \mathbb{R}^{N_q}$.
\begin{equation}
    \bm{A} = \text{FC}([\bm{D}_A; {\bm{F}}_M]).
\end{equation}

\subsection{Loss Function}

Followed by the query-based object detection method ~\cite{kuhn1955hungarian,liao2022gen,wang2024ted}, the matching loss is designed in IIM between the predicted interaction pairs $\bm{Y}=[\bm{H}_B, \bm{O}_B, \bm{O}_C, \bm{A}]$ and the ground-truth pairs $ \bm{GT}_{pair} $, as follows:
\begin{equation}
    \setlength{\abovedisplayskip}{2pt}
    \setlength{\belowdisplayskip}{-3pt}
    \mathcal{L}_{m\_l} = \sum_{i}^{N_q}\mathcal{L}_{m}(\bm{GT}_{pair}^i, \bm{Y}^i), 
\end{equation} 
\begin{equation}
    \mathcal{L}_{m}= \sum_{p\in \bm{O}_C,\bm{A}} \mathcal{L}_{cls}^p+ \sum_{q\in \bm{H}_B,\bm{O}_B} \mathcal{L}_{box}^q + \sum_{r\in \bm{H}_B,\bm{O}_B} \mathcal{L}_{iou}^r.
\end{equation}
For each sample, the total matching loss $\mathcal{L}_{\text{m\_l}}$ is defined as the sum of individual matching losses $\mathcal{L}_{\text{m}}$. Specifically, $\mathcal{L}_{\text{m}}$ consists of: (1) classification loss $\mathcal{L}_{\text{cls}}$ for accurate interaction recognition; (2) bounding box regression loss $\mathcal{L}_{\text{box}}$ for localization refinement; and (3) IoU loss $\mathcal{L}_{\text{iou}}$ to further improve prediction-ground truth alignment. The total loss $\mathcal{L}_{t}$ is computed via:
\begin{equation}
\label{eq:total_loss}
    \mathcal{L}_{t} = \alpha\mathcal{L}_{m\_l} + \beta(\mathcal{L}_{BCE}(\bm{GT}_L, \bm{L}) + \mathcal{L}_{CE}(\bm{GT}_s, \bm{S})),
\end{equation}
where BCE measures the contact prediction $\bm{L}$ against $\bm{GT}_L$. The Cross Entropy (CE) loss evaluates the accuracy of $\bm{S}$ compared to $\bm{GT}_S$. Hyperparameters $\alpha$ and $\beta$ balance the segment and interaction tasks.

\begin{table*}[h]
\centering

\setlength{\tabcolsep}{1.4mm}{
\begin{tabular}{lll|ccccc}
\specialrule{1.5pt}{0pt}{0pt} 
\multirow{2}{*}{Methods}                                                                                                       & \multirow{2}{*}{Params} & \multirow{2}{*}{Backbone} & \multicolumn{1}{c}{\multirow{2}{*}{mAP$\uparrow$}} & \multirow{2}{*}{SC-Acc.$\uparrow$} & \multirow{2}{*}{C-Acc.$\uparrow$} & \multirow{2}{*}{mIoU$\uparrow$} & \multirow{2}{*}{wIoU$\uparrow$} \\
                                                                                                         &                      &                           & \multicolumn{1}{c}{}                     &                          &                         &                       &                       \\ \hline
Trans~\shortcite{tamura2021qpic}+UNet~\shortcite{ronneberger2015u}   & 74.0M                  & ResNet-50                 & 47.93                                     & 12.64                    & 24.99                   & 7.84                 & 13.65               \\
RCLTrans~\shortcite{kim2023relational}+UNet~\shortcite{ronneberger2015u}    & 101.5M           & ResNet-50                 & 52.89                                    & 12.93                    & 26.24                   & 7.96                  & 14.04                \\

DisTrans~\shortcite{wang2024ted}+UNet~\shortcite{ronneberger2015u}       & 79.6M                                                       & ResNet-50                 & 53.59                                    & 12.52 & 24.79 & 8.30 & 13.79               \\
DisTrans~\shortcite{wang2024ted}+LinkNet~\shortcite{chaurasia2017linknet}  & 120.2M                                                     & ResNet-50                 & 53.17                                    & 14.98 & 28.25 & 10.74 & 18.00                  \\
DisTrans~\shortcite{wang2024ted}+MANet~\shortcite{fan2020ma}     & 236.5M                                                               & ResNet-50                 & 53.03                                    & 15.78 & 31.76 & 10.81 & 17.45                 \\
DisTrans~\shortcite{wang2024ted}+HOT~\shortcite{hot}         & 97.3M                                                                   & ResNet-50                 & 53.45                                   &  17.45 & 32.24 & 11.54 & 16.56                \\
DisTrans~\shortcite{wang2024ted}+PIHOT~\shortcite{wang2025precision} & 158.4M                                                                        & ResNet-50                 & 53.71                                    &  17.03                        &  35.87                       &  11.98                     &  18.30                     \\
\rowcolor[gray]{0.9}Ours     & \textbf{55.5M}                                                                                                                      & ResNet-50                 & \textbf{57.60}                                    & \textbf{19.80}                    & \textbf{42.40}                   & \textbf{13.25}                 & \textbf{20.37}                 \\
\rowcolor[gray]{0.9}Ours     & 74.4M                                                                                                                      & ResNet-101                & 59.06                     & 21.30  & 45.42 & 14.11 & 21.50                               \\ 
\rowcolor[gray]{0.9}Ours     & 73.5M                                                                                                                      & Swin-S                & 59.43                                    & 22.41                        & 47.30                       & 15.97                      &  22.31                     \\ 
\rowcolor[gray]{0.9}Ours     & 224.1M                                                                                                                      & Swin-L                & \textbf{61.32}                                    & \textbf{24.01}                        & \textbf{49.62}                       &   \textbf{17.11}                    &  \textbf{23.79}                     \\ 
\specialrule{1.5pt}{0pt}{0pt} 
\end{tabular}}
\caption{Performance comparisons on PaIR-2 dataset. }
\label{tab:pair-2}
\end{table*}

\begin{table*}[t]
  \centering
  \begin{minipage}[t]{0.48\linewidth}
    \centering
    
    \setlength{\tabcolsep}{1.0mm}{
    \begin{tabular}{l|ccccc}
\specialrule{1.5pt}{0pt}{0pt}
\multirow{2}{*}{Module} & \multicolumn{1}{c}{\multirow{2}{*}{mAP}} & \multirow{2}{*}{SC-Acc.} & \multirow{2}{*}{C-Acc.} & \multirow{2}{*}{mIoU} & \multirow{2}{*}{wIoU} \\
                        & \multicolumn{1}{c}{}                     &                          &                         &                       &                       \\ \hline
baseline                & 32.08                                     & 23.80                    & 43.66                   & 13.33                  & 17.10                 \\
+CPAM                     & 32.27                                    & 26.04                    & 46.16                   & 15.94                  & 19.16                 \\
+H-O              & 32.63                                    & 29.21                    & 50.01                   & 17.37                 & 21.92                 \\
+M-G              & \textbf{35.09}  & \textbf{29.93}  & \textbf{50.87}  & \textbf{17.79}  & \textbf{22.23}               \\ \specialrule{1.5pt}{0pt}{0pt}
\end{tabular}}

    \caption{Performance of each component when using ResNet-50 backbone. 
    The \textbf{+} indicates incremental addition of modules. The H-O and M-G denote the H-O RoI enhancer and mask-guided ROI feature modules, respectively.}
    \label{tab:ab_modules}
    
  \end{minipage}
  \hfill
  \begin{minipage}[t]{0.48\linewidth} 
    \centering
    
    \setlength{\tabcolsep}{1.0mm}{
\begin{tabular}{ll|ccccc}
\specialrule{1.5pt}{0pt}{0pt}
\multirow{2}{*}{$\alpha$} & \multirow{2}{*}{$\beta$} & \multicolumn{1}{c}{\multirow{2}{*}{mAP}} & \multirow{2}{*}{SC-Acc.} & \multirow{2}{*}{C-Acc.} & \multirow{2}{*}{mIoU} & \multirow{2}{*}{wIoU} \\
                                       &                                       & \multicolumn{1}{c}{}                     &                          &                         &                       &                       \\ \hline
0.1                                    & 1.0                                   & 34.59                                & 29.02                   & 50.43                  & 17.45  & 21.96               \\
0.5                                    & 1.0                                   & 34.38 & 28.41 & 49.84 & 17.24 & 21.73                 \\
1.0                                    & 1.0                                   & 34.06                                    &  28.48                   & 48.36                   & 16.96                 & 21.25                 \\
0.1                                    & 0.5                                   & \textbf{35.09}                                    & \textbf{29.93}                    & \textbf{50.87}                   & \textbf{17.79}                 & \textbf{22.23}                 \\
0.5                                    & 0.1                                   & 33.81                                    & 27.78                    & 48.09                   & 16.35                 & 20.97                 \\
1.0                                    & 0.1                                   & 33.39                                    & 27.09                    & 48.53                   & 16.02                 & 20.19                 \\ \specialrule{1.5pt}{0pt}{0pt}
    \end{tabular}}
  \caption{The impact of different values of \( \alpha \) and \( \beta \) in Eq. \ref{eq:total_loss} on the experimental results when using ResNet-50 backbone.}
    \label{tab:ab_loss}  
  \end{minipage}
  
\end{table*}

\section{Experiment}
\label{sec:experiment}
\subsection{Datasets}

To advance research on interaction and fine-grained contact region segmentation, we introduce a high-quality dataset unifying interaction pair recognition and pixel-level contact annotation in real-world images. We integrate diverse sources from COCO~\cite{lin2014microsoft}, HICO~\cite{chao2015hico}, HAKE~\cite{li2020pastanet}, Watch-n-Patch~\cite{wu2015watch}, and task-specific datasets like HICO-Det~\cite{chao2018learning}, V-COCO~\cite{gupta2015visual}, and HOT~\cite{hot,wang2025precision}. Through cross-source integration and re-annotation, we build a dataset with strict interaction semantics and high-quality contact segmentation labels. 
%

Our dataset, PaIR, is the first to establish pixel-level contact annotation guided by interaction semantics in real-world images. After reviewing tens of thousands of interaction instances across nearly 100,000 images, we selected 13,979 images for annotation. The final dataset includes \textbf{45,103} instances, \textbf{46,616} interaction pairs, and \textbf{32,301} contact regions, covering \textbf{654} action categories and \textbf{80} object categories. The dataset is divided into two subsets: \textbf{PaIR-1} (8,591 images, 30,309 instances, 22,896 interaction pairs, 21,312 contact regions, 430 action categories) and \textbf{PaIR-2} (5,388 images, 14,794 instances, 23,720 interaction pairs, 10,989 contact regions, 224 action categories). PaIR-1 covers diverse action categories with broader interaction range and more challenging tasks, while PaIR-2 focuses on common daily object interactions, suitable for evaluating models in general-purpose scenarios.

\subsection{Implementation Details}
\label{sec:exp_imple}
The weights of CPAM and PGCS are initialized using the method proposed by He et al.~\cite{he2015delving}. For the encoder part of the IIM, we directly load the pre-trained DETR~\cite{he2015delving}. The AdamW optimizer is used to optimize the network with an initial learning rate of \( 1 \times 10^{-4} \). In the loss function, the weight coefficients are set to \(\alpha = 0.1\) and \(\beta = 0.5\). 
The entire model is trained on 8 NVIDIA A6000 (48G) GPUs, with a batch size of 4 per GPU. Our implementation is based on PyTorch 1.7.1 and TorchVision 0.8.2, running on Ubuntu 22.04. 

\subsection{Evaluation Metric}
For contact region segmentation evaluation, we adopt four metrics proposed by Chen et al.~\cite{hot}: SC-Acc., C-Acc., mIoU, and wIoU. SC-Acc. measures the proportion of pixels correctly identified as ``in contact'' and correctly associated with corresponding body part labels. C-Acc. evaluates binary classification accuracy at pixel level for contact. mIoU is the average IoU across all categories, while wIoU is the class-weighted average IoU based on pixel proportions. For interaction detection, we adopt mAP as the evaluation criterion~\cite{wang2024ted}.

\begin{figure*}[h]
\centering 
\includegraphics[width=\linewidth]{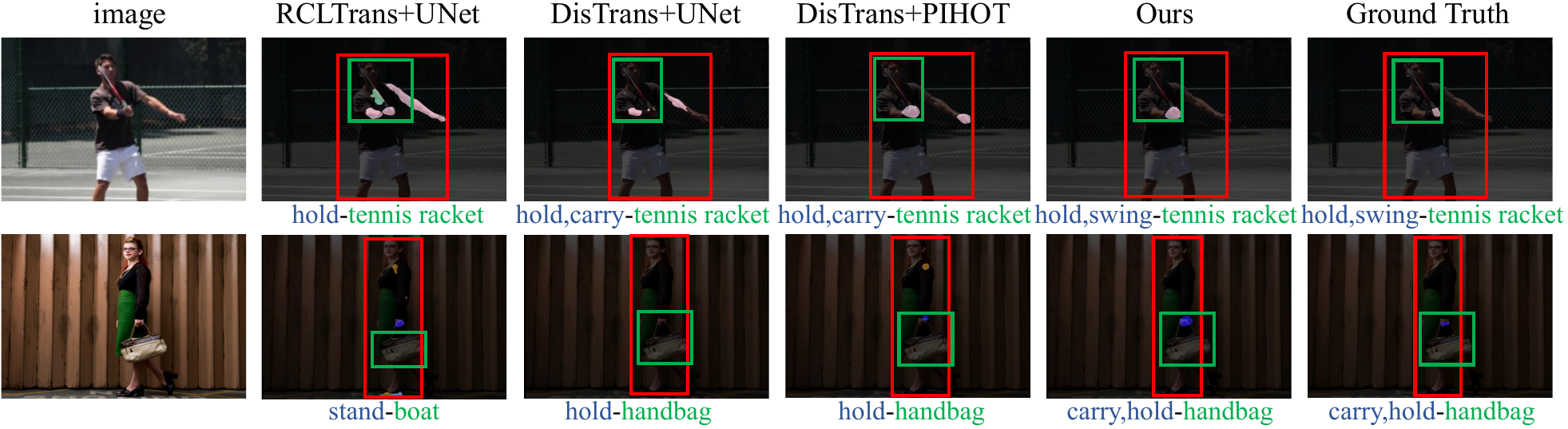}
\caption{Visualization results. Red and green bounding boxes represent the human and object, respectively. 
Blue text indicates the action category, and green text indicates the object category.
}
\label{fig:vis_results}
\end{figure*}

\begin{table*}[t]
\small
  \centering
  \begin{minipage}[t]{0.48\linewidth}
    \centering
    
    \setlength{\tabcolsep}{0.8mm}{
    \begin{tabular}{l|ccccc}
\specialrule{1.5pt}{0pt}{0pt}
\multirow{2}{*}{Module} & \multirow{2}{*}{mAP} & \multirow{2}{*}{SC-Acc.} & \multirow{2}{*}{C-Acc.} & \multirow{2}{*}{mIoU} & \multirow{2}{*}{wIoU} \\
                        &                      &                          &                         &                       &                       \\ \hline
Trans~\shortcite{tamura2021qpic}                   & 13.70                 & 29.31                    & 50.19                   & 17.57                  & 21.98                 \\
RCLTrans~\shortcite{kim2023relational}                & 25.85               & 29.17                    & 51.01                   & 17.59                 & 22.04                 \\
STTrans~\shortcite{zhang2022exploring}            & 30.21                & 27.16                    & 49.85                   & 17.33                  & 22.17                 \\
DisTrans~\shortcite{wang2024ted}                &  31.92                    & 29.76                         & 50.74                        &   17.52                    & 21.88                      \\
IIM                      & \textbf{35.09}  & \textbf{29.93}  & \textbf{50.87}  & \textbf{17.79}  & \textbf{22.23}                \\ \specialrule{1.5pt}{0pt}{0pt}
\end{tabular}}
\caption{Experimental results of replacing different interaction detection modules when using ResNet-50 backbone. Note that only the proposed IIM is replaced, while all other components remain unchanged.}
    \label{tab:ab_ar_replace}
  \end{minipage}
  \hfill
  \begin{minipage}[t]{0.48\linewidth} 
    \centering
    
    \setlength{\tabcolsep}{1.0mm}{
    \begin{tabular}{l|ccccc}
\specialrule{1.5pt}{0pt}{0pt}
\multirow{2}{*}{Module} & \multirow{2}{*}{mAP} & \multirow{2}{*}{SC-Acc.} & \multirow{2}{*}{C-Acc.} & \multirow{2}{*}{mIoU} & \multirow{2}{*}{wIoU} \\
                        &                      &                          &                         &                       &                       \\ \hline
UNet~\shortcite{ronneberger2015u}                    & 34.85                 & 15.67                    & 27.15                   & 9.93                  & 16.18                 \\
LinkNet~\shortcite{chaurasia2017linknet}                 & 34.52                & 22.02                    & 36.91                   & 12.85                  & 16.95                 \\
MANet~\shortcite{fan2020ma}                   & 35.01                & 24.29                    & 46.18                   & 13.41                & 17.99                 \\
HOT~\shortcite{hot}                     &  34.78                    & 24.86                         & 41.63                        &  14.04                     &  18.70                     \\
PIHOT~\shortcite{wang2025precision}                   & 34.69                & 26.89                    & 47.58                   & 15.18                 & 19.74                 \\
PGCS                      & \textbf{35.09}  & \textbf{29.93}  & \textbf{50.87}  & \textbf{17.79}  & \textbf{22.23}  \\ \specialrule{1.5pt}{0pt}{0pt}
\end{tabular}}
    \caption{Performance of replacing different contact segmentation modules when using ResNet-50 backbone. All other components of the proposed framework remain unchanged.}
    \label{tab:ab_cs_replace}
  \end{minipage}
\end{table*}

\subsection{Results}
\label{sec:exp_results}

We evaluate our method on both PaIR-1 and PaIR-2 datasets. As shown in Table~\ref{tab:pair-1}, on PaIR-1, we compare against some interaction detection methods (Trans~\cite{tamura2021qpic}, RCLTrans~\cite{kim2023relational}, STTrans~\cite{zhang2022exploring}, DisTrans~\cite{wang2024ted}) and contact segmentation models (UNet~\cite{ronneberger2015u}, LinkNet~\cite{chaurasia2017linknet}, MANet~\cite{fan2020ma}, HOT~\cite{hot}, PIHOT~\cite{wang2025precision}). 
With ResNet-50 backbone, our approach achieves best performance across all metrics—mAP (35.09), SC-Acc. (29.93), C-Acc. (50.87), mIoU (17.79), and wIoU (22.23)—surpassing the second-best method by a notable margin. \textbf{Moreover, our model maintains a compact size of only 55.6M parameters, significantly fewer than comparable methods, and achieves the fastest inference time of 75.2 ms, demonstrating superior efficiency-performance trade-off.
} Performance further improves with stronger backbones like ResNet-101, Swin-S~\cite{liu2021swin}, and Swin-L~\cite{liu2021swin}.

Table~\ref{tab:pair-2} presents results on PaIR-2, where our method again outperforms all baselines across all evaluation metrics using ResNet-50, while retaining the lowest parameter count (55.5M), reinforcing its effectiveness and efficiency.

\subsection{Ablation Study}

To evaluate the contribution of each proposed component, we conduct an ablation study on PaIR-1, with results summarized in Table~\ref{tab:ab_modules}. 
The baseline configuration includes only the PGCS module (without the H-O RoI Enhancer) and the interaction reasoning module (without the Mask-Guided RoI Feature). Under this setup, the model achieves 32.08 mAP, 23.80 SC-Acc., 43.66 C-Acc., 13.33 mIoU, and 17.10 wIoU. 
Incorporating the CPAM module yields noticeable gains in contact segmentation by enhancing the PGCS branch. Further addition of the H-O RoI Enhancer (+H-O) leads to improved contact performance, achieving 29.21, 50.01, 17.37, and 21.92 across the respective metrics. Finally, integrating the Mask-Guided RoI Feature (+M-G) substantially enhances interaction detection, resulting in the highest overall performance across all evaluation criteria.

We further analyze the sensitivity of the model to the hyperparameters \( \alpha \) and \( \beta \) in Eq. \ref{eq:total_loss}, as shown in Table~\ref{tab:ab_loss}. The best performance is achieved when \(\alpha = \text{0.1}\) and \(\beta = \text{0.5}\). 
Notably, \(\alpha\) is smaller than \(\beta\), primarily because the loss computed for interaction detection tends to be larger than that for segmentation. 
Therefore, to maintain a balanced contribution between the two tasks, it is necessary to assign a smaller weight to \(\alpha\) than to \(\beta\).

We also evaluate the effect of replacing the proposed IIM and PGCS modules with existing methods. As IIM and PGCS are responsible for interaction detection and contact segmentation, respectively, substituting IIM primarily affects detection performance, while replacing PGCS impacts segmentation. Table~\ref{tab:ab_ar_replace} reports the results of using alternative interaction detectors. As detector capability improves, overall performance increases. Similarly, Table~\ref{tab:ab_cs_replace} shows that our PGCS module outperforms other methods, benefiting from the integration of interaction context to improve contact region discrimination.

\subsection{Visualization}
Figure~\ref{fig:vis_results} presents qualitative results of our method. The predicted contact regions are highlighted against a black background. The first row illustrates the action ``hold and swing tennis racket'', where our approach accurately identifies all interaction instances and correctly segments the contact between the right hand and the racket. In contrast, alternative methods exhibit failures in interaction recognition or contact segmentation. Similar observations in the second row further demonstrate the effectiveness of our method in both interaction understanding and contact localization.

\section{Conclusion} In this paper, we propose PaIR-Net, a novel and unified framework that jointly performs interaction recognition and pixel-level segmentation of contact regions. To enable effective joint modeling, we introduce several complementary and synergistic modules—CPAM, PGCS, and IIM—which collaboratively fuse high-level interaction semantics with fine-grained structural contact cues. To facilitate research in this domain, we also construct a high-quality composite dataset that, for the first time, provides joint annotations of interacting object pairs and their corresponding contact regions in real-world scenarios. Extensive experiments on benchmark datasets show that our method consistently outperforms state-of-the-art approaches across multiple evaluation metrics, highlighting the effectiveness and robustness of our proposed framework.

\section{Acknowledgments}
This work was supported in part by the National Natural Science Foundation of China under Grant 62202174, in part by the GJYC program of Guangzhou under Grant 2024D01J0081, and in part by the ZJ program of Guangdong under Grant 2023QN10X455, and in part by the Fundamental Research Funds for the Central Universities under Grant 2025ZYGXZR053.

\bibliography{aaai2026}

\end{document}